\documentclass[letterpaper, 10 pt, conference]{ieeeconf}  

\IEEEoverridecommandlockouts                              

\usepackage{graphics} 
\usepackage{epsfig} 
\usepackage{times} 
\usepackage{amsmath} 
\usepackage{mathrsfs}
\usepackage{bm}
\usepackage{stfloats}
\usepackage{amssymb}  
\usepackage{subfigure}
\usepackage[T1]{fontenc}
\usepackage{booktabs}
\usepackage{multirow}

\usepackage{graphicx}
\usepackage{float}

\usepackage{cite}
\usepackage{xcolor}
\usepackage{url}
\makeatletter
\let\NAT@parse\undefined
\makeatother
\usepackage{hyperref}
\usepackage{algpseudocode}
\usepackage[export]{adjustbox}
\usepackage{balance}
\usepackage[utf8]{inputenc}
\usepackage{verbatim}
\usepackage{makecell}

\renewcommand{\theenumii}{\theenumi.\arabic{enumii}.}
\usepackage{algorithm}  
\usepackage{algorithmicx}
\usepackage{algpseudocode}  
\usepackage{amsmath}

\usepackage{etoolbox}
\makeatletter
\patchcmd{\@makecaption}
{\scshape}
{}
{}
{}
\makeatother

\title{\LARGE \bf
FAEP: Fast Autonomous Exploration Planner for UAV Equipped with Limited FOV Sensor
}
\author{ Yinghao Zhao$^{1}$, Li Yan$^{1}$, Yu Chen$^{2}$, Hong Xie$^{1}$,  Bo Xu$^{1}$
\thanks{$^{1}$Yinghao Zhao, Li Yan,  Hong Xie and  Bo Xu are with School of Geodesy and Geomatics, Wuhan University, Wuhan 430079, China; Corresponding author: Li Yan }
\thanks{
$^{2}$Yu Chen is with School of Geomatics Science and Technology, Nanjing Tech University, Nanjing 211816, China}
}

\begin{document}
    \maketitle
    \pagestyle{empty}  
    \thispagestyle{empty} 
   \thispagestyle{empty}
	\pagestyle{empty}

\begin{abstract}
Autonomous exploration is one of the important parts to achieve the autonomous operation of Unmanned Aerial Vehicles (UAVs). To improve the efficiency of the exploration process, a fast and autonomous exploration planner (FAEP) is proposed in this paper. We firstly design a novel frontiers exploration sequence generation method to obtain a more reasonable exploration path, which considers not only the flight-level but frontier-level factors into TSP. According to the exploration sequence and the distribution of frontiers, a two-stage heading planning strategy is proposed to cover more frontiers by heading change during an exploration journey. To improve the stability of path searching, a guided kinodynamic path searching based on a guiding path is devised. In addition, a dynamic start point selection method for replanning is also adopted to increase the fluency of flight. We present sufficient benchmark and real-world experiments. Experimental results show the superiority of the proposed exploration planner compared with typical and state-of-the-art methods.
\end{abstract}

\section{INTRODUCTION}
\label{sec:introduction}
UAV has been widely used in surveying and mapping\cite{qin2019autonomous, battulwar2020practical, petravcek2021large}, environmental protection, rescue, military, and other fields in recent years due to its unique advantages, and remarkable results have been achieved. However, in most operation scenarios, it is still in the state of human operation, and there is still a long way to go in autonomous operation ability. As one of the key parts of UAV autonomous capability, autonomous exploration has attracted extensive attention in recent years, and many excellent autonomous exploration algorithms have emerged \cite{bircher2016receding, meng2017two,cieslewski2017rapid,selin2019efficient,dang2020graph,dharmadhikari2020motion }.

Although existing robot autonomous exploration methods can explore environments by using frontiers or sampling viewpoints, there are still many problems to be solved. The methods using frontiers can quickly explore the whole environment by searching frontiers and generating an exploration sequence, but the process of finding and describing frontiers is always computationally expensive. Although the methods by sampling viewpoints can easily generate the candidate goals, it always causes a low exploration rate and efficiency. In addition, most of the existing methods are using greed strategy, which pays attention to the local information gain but ignores the global exploration efficiency. And few algorithms consider the dynamics of UAV, which will cause unsmooth exploration trajectory, low-speed flight, and lots of stop-and-go maneuvers. FUEL \cite{zhou2021fuel} is a state-of-the-art fast and autonomous exploration algorithm. Its heuristic framework can achieve rapid and efficient UAV exploration in complex environments through the designed incremental frontier structure (FIS) and hierarchical planning. And it can generate smooth and high-speed exploration trajectory in high frequency. However, although this algorithm has greatly improved the exploration rate and exploration efficiency compared with other algorithms, it still faces problems affecting its exploration efficiency, such as back-and-forth maneuvers during the exploration process.

To solve the above problems, based on the framework of FUEL, this paper proposes a fast and autonomous UAV exploration algorithm (FAEP). In the part of global exploration path generation, the influence of frontier-level on global exploration is considered, and a corresponding quantitative method is designed. By combining it with flight-level factors, a better frontiers exploration sequence with a low proportion of back-and-forth maneuvers is proposed. After the next exploration target is determined, a two-stage heading planning method is designed to achieve more efficient exploration by covering more frontiers through heading change in one flight. And then, in order to improve the stability of path planning and avoid the problem of search failure or time-consuming in some special environments, guided kinodynamic path searching is designed, which uses a geometric path to guide the direction of kinodynamic path searching. In addition, a dynamic start point selection method for replanning is also adopted to increase the fluency of flight.

We compare our method with three typical and state-of-the-art methods in different simulation environments. The experimental results show that our method and FUEL have obvious advantages over the other two methods, and the exploration speed is 3-6 times faster. Compared with FUEL, the exploration time of our method in two different environments 
is shortened by 28.7 $\%$ and 12.8$\%$ , and the exploration path is shortened by 26.3$\%$  and 11.2$\%$  respectively. In addition, we also verify the effectiveness of our method through onboard real-world exploration. The contributions of this paper are as follows:
 
\begin{itemize}
	\item A better frontier exploration sequence generation method, which considers not only flight-level but frontier-level factors to generate a more reasonable global path.
	\item A two-stage heading planning method for covering more frontiers when flying to the viewpoint.
	\item A guided kinodynamic path searching method based on the guiding path and a dynamic planning strategy, which improves the stability and fluency of the flight.
	\item Simulation and real-world experiments are carried out in various environments.

\end{itemize}

\section{RELATED WORK}\label{sec:related work}  
The problem of autonomous exploration has been studied by many scholars in recent years, and lots of methods from multiple angles have been proposed, which are mainly divided into the following three categories: sampling-based exploration \cite{connolly1985determination,cao2021tare,zhu2021dsvp,   wang2019efficient,witting2018history,  dang2018visual, oleynikova2018safe,xu2021autonomous,respall2021fast }, frontier-based exploration \cite{yamauchi1997frontier,shen2012stochastic,  deng2020robotic, heng2015efficient, zhong2021information,julia2012comparison, dai2020fast, batinovic2021multi } and algorithms based on machine learning which has emerged recently \cite{maciel2019online}. This paper only discusses the previous two algorithms which have been widely used in various exploration tasks.

Sample-based exploration methods use randomly sampled viewpoints in the free space, which find the next best view by obtaining a path with the highest information gain. A receding horizon “next-best-view” scheme (NBVP) is proposed to explore the 3D environments by considering the information gained over the entire path in \cite{bircher2016receding}. NBVP is the first method that uses the concept of the next best view for exploration in a receding horizon fashion, and many methods are derived from this method. These methods select the path with the highest information gain in the incrementally RRT for UAVs to execute. The method (Aeplanner) in \cite{selin2019efficient} combines frontier exploration and NBVP to avoid getting stuck in large environments not exploring all regions, and the method also makes the process of estimating potential information gain faster by using cached points from earlier iterations. An incremental sampling and probabilistic roadmap are used in \cite{xu2021autonomous} to improve the efficiency of planning. The method \cite{respall2021fast} uses a combination of sampling and frontier-based method to reduce the impact of finding unexplored areas in large scenarios. There are also some two-stage methods \cite{cao2021tare,zhu2021dsvp} to cover the entire environment efficiently by different planning strategies in the global and local map.
x
In contrast, the frontier-based method is mainly comprised of two processes, finding frontiers (the boundary between mapped and unmapped areas) and solving a sequence problem for a global path to visit frontiers. The first frontier-based exploration method is introduced by \cite{yamauchi1997frontier} to explore a generic 2D environment, which selects the closest frontier as the next goal. And then, a stochastic differential equation-based exploration algorithm \cite{shen2012stochastic} is proposed to achieve exploration in 3D environments. To achieve high-speed flight, \cite{cieslewski2017rapid} proposed a method that extracts frontiers in the field of view (FOV) and selects the frontier minimizing the velocity change. For finding a reasonable frontier exploration sequence, the traveling salesman problem (TSP) is employed in \cite{meng2017two}. A wise exploration goal is selected by adopting an information-driven exploration strategy in \cite{zhong2021information}. However, many methods are facing the problems of inefficient global coverage, conservative flight trajectory, and low decision frequencies. For solving these issues,  \cite{zhou2021fuel} achieved fast exploration in complex environments by adopting an incremental frontier structure and hierarchical planning. This method not only generates a high-quality global exploration sequence but also generates a fast and smooth flight trajectory in a short time. And this paper is an extension of the framework in \cite{zhou2021fuel}.
\begin{figure}[htpb]
\vspace{2mm}
\centering
\includegraphics[width=.9\linewidth]{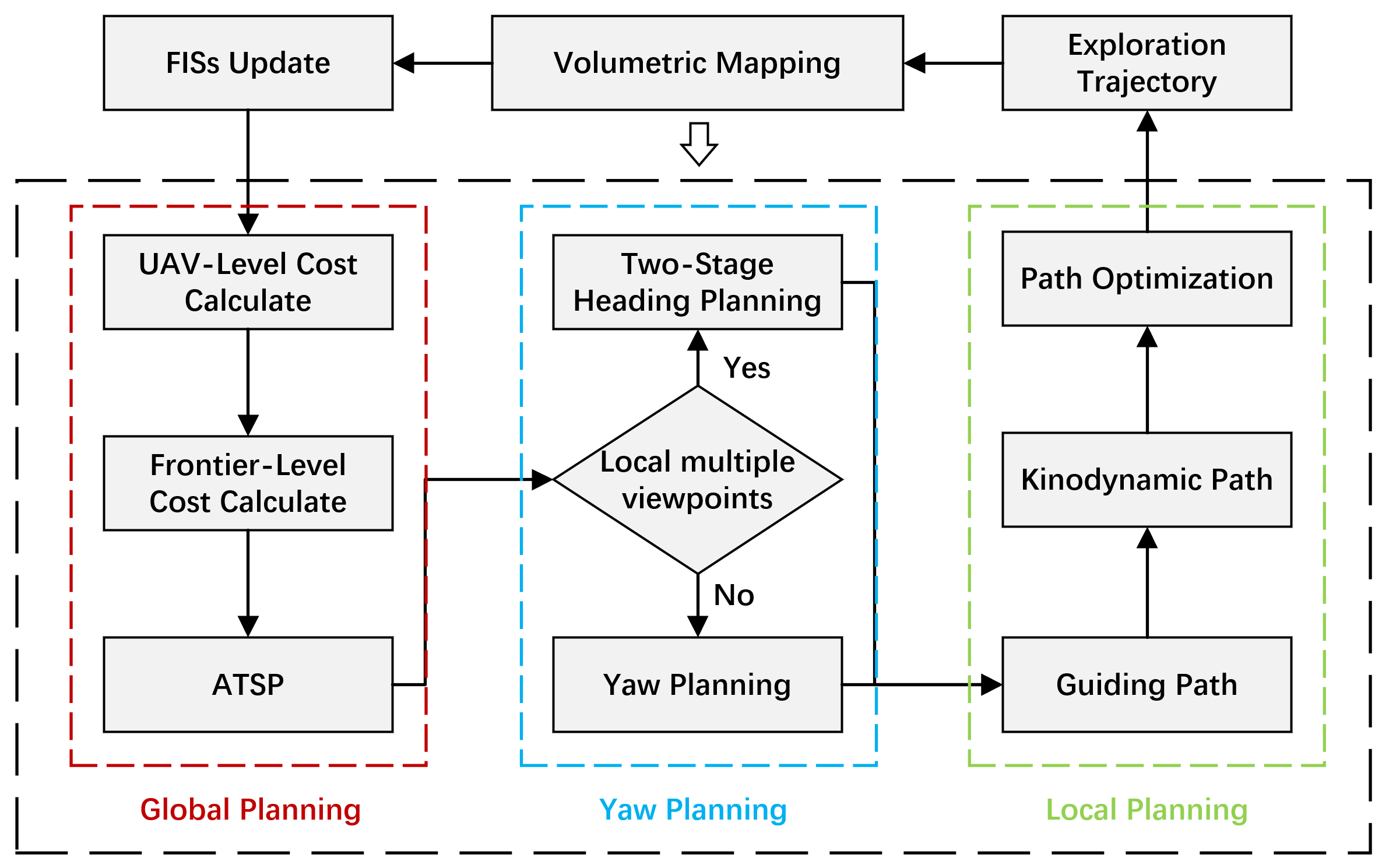}
\caption{An overview of the proposed fast autonomous exploration planner}
\label{fig:1}
\vspace{-3mm}
\end{figure}\textbf{}
\section{PROPOSED APPROACH}\label{sec:proposed approach}
Our method is improved on the basis of FUEL \cite{zhou2021fuel}. The main operation flow is shown in Fig.\ref{fig:1}. After lots of exploration tasks with FUEL, we observe that there are some back-and-forth and stop-and-go maneuvers causing the decline of efficiency during the exploration process, which is due to the low quality of frontiers exploration sequence and instability of path searching in individual environments. In order to reduce the occurrence of the above two situations, we design a frontiers exploration sequence generation method considering the global exploration influence of frontiers and a guided exploration path planning method. Meanwhile, to achieve more efficient exploration, this paper designs a two-stage heading planning method based on the distribution of frontiers, which covers more frontiers in the process of flying to the viewpoint. In addition, we also adopt a dynamic start point for exploration replanning to improve the fluency of flight.

\subsection{Better Frontiers Exploration Sequence}
The frontiers exploration sequence is crucial for the frontier-based exploration method. The rationality of the frontiers exploration sequence directly affects the efficiency of the whole exploration process. Many methods use TSP to obtain the best exploration sequence. However, most methods only take the Euclidean distance between the frontiers as the cost of TSP, which is simple but obviously insufficient. FUEL does not use the conventional TSP but uses a more reasonable ATSP for the solution, and it not only takes the Euclidean distance as the cost but also takes the yaw change and speed direction change in the next stage as one of the costs to generate an exploration sequence. It optimizes the exploration sequence to a certain extent, but there are still some deficiencies. The factors considered only stay at the flight-level (flight distance, yaw change, speed change), and do not consider the frontier-level for global rapid exploration. This often cause the back-and-forth maneuvers, which will increase the exploration time and flight distance. 

In order to overcome the above shortcomings, a more reasonable frontiers exploration sequence is proposed. Compared with other exploration algorithms that only consider the factors of the current flight-level, this method also considers the influence of the frontier on global exploration. This paper holds that when the frontier is an independent small area or a frontier close to the boundary of the exploration area, the corresponding exploration priority should be higher. If this area is not preferentially explored, it will lead to back-and-forth maneuvers and reduce the efficiency of global exploration. To solve this, this paper designs two principles: edge priority principle and independent small area priority principle.

To achieve the edge priority, we calculate the shortest distance $d_{kmin}$ between average point $p_{ka}$ of FIS $F_{k}$ in FISs and the boundary of the exploration area: 
\begin{flalign}
\begin{aligned}
    d_{kmin} = min(d_{kx}, d_{ky},d_{kz}) 
\end{aligned}\label{equ:1}
\end{flalign}
where $d_{kx}$,  $d_{ky}$,  $d_{kz}$ is the shortest distance from X, Y, Z-axis. We regard $d_{kmin}$ as one of the frontier costs in ATSP to obtain a sequence where the frontiers near the exploration boundary will be explored in priority. In order to maintain the the efficiency of the principle, we assume that the range of the exploration area is boundaried by a box $(B_x, B_y, B_z)$. When there is a range less than $B_{min}(15, 15, 10)$, we remove the axis from Equ.\ref{equ:1}. And $B_{min}$ depends on the maximum range of the sensor on each axis. Here, we choose three times of maximum range of the sensor to generate $B_{min}$.
\begin{figure}[t]
	\centering
	\vspace{2mm}
	\includegraphics[width=.8\linewidth]{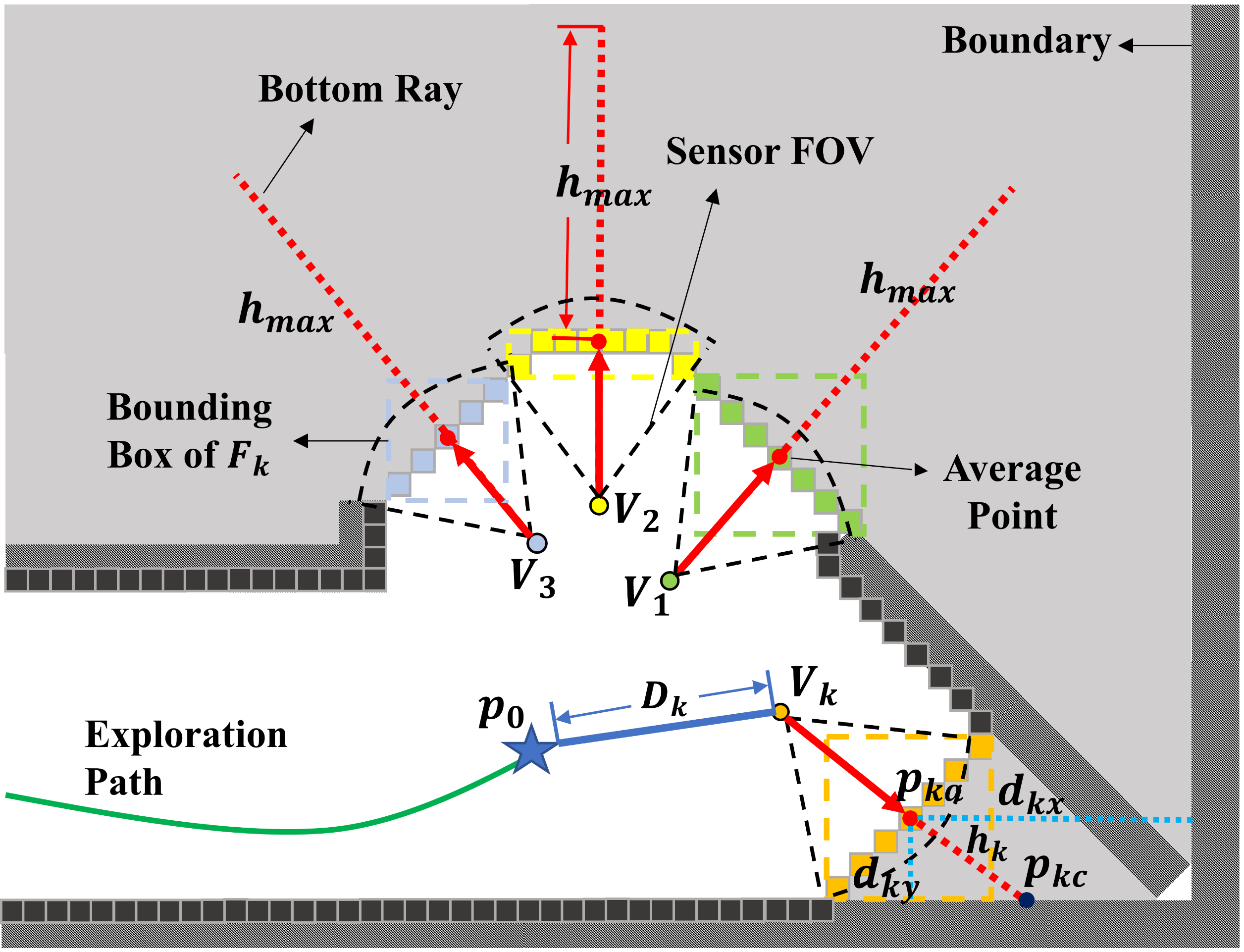}

	\caption{The proposed two-level frontiers cost calculation method for generating a better frontiers exploration sequence.}
	\label{fig:2}
	\vspace{-6mm}
\end{figure}\textbf{}

To achieve the independent small area priority principle, a method called Bottom Ray is designed as shown in Fig.\ref{fig:2}. Firstly, we obtain the viewpoints that the distance between the viewpoints $V_k (p_k,  \xi_k)$ and the current position $p_0$ of UAV is less than $D_{thr}$. Each viewpoint $V_k$ contains a position  $p_k$ and a yaw angle $\xi_k$. Secondly, the vector $\overrightarrow{p_kp_{ka}}$  from the position  $p_k$ of the viewpoint to the average point in FIS $F_k$ is calculated. Thirdly, extending the vector according to the mapping resolution until it touches the occupied, free voxel, boundary or exceeds the set maximum distance $h_{max}$, then a bottom point $p_{kc}$ is obtained, and we regard the distance $h_k$ between $p_{ka}$ and $p_{kc}$ as the influence of the frontier on global exploration. Finally, we regard $h_k$ and $d_{kmin}$ as the factors of  frontier-level, and integrate flight-level factors used in FUEL and the frontier-level factors into the cost matrix $\mathrm{M_{tsp}}$ of ATSP as follows: 
\begin{flalign}
	\begin{aligned}
		&\mathrm{M}_{\mathrm{tsp}}(0, k)=t_{\mathrm{lb}}\left(V_{0}, 
		~V_{k}\right)+w_{{c}} \cdot c_{{c}}\left(V_{k}\right) \\
		&\quad+w_{{b}} \cdot d_{kmin}-w_{{f}} \cdot\left(h_{\max }-h_{k}\right) \\
		&k \in\left\{1,2, \cdots, N_{\mathrm{cls}}\right\}
	\end{aligned}   \label{equ:2}
\end{flalign}
\begin{flalign}
	\begin{aligned}
		&t_{\mathrm{lb}}\left(V_{0}, ~V_{k}\right)=\max 
		\left\{\frac{\operatorname{length}\left(P\left(p_{0}, 
			p_{k}\right)\right)}{v_{\max }}\right. , \\
		&\left.\frac{\min \left(\left|\xi_{0}-\xi_{k}\right|, 2 
			\pi-\left|\xi_{0}-\xi_{k}\right|\right)}{\dot{\xi}_{\max }}\right\}
	\end{aligned}
\end{flalign}
\begin{flalign}
	c_{c}\left(V_{k}\right)=\cos ^{-1} \frac{\left(p_{k}-p_{0}\right) \cdot 
		v_{0}}{\left\|p_{k}-{p}_{0}\right\|\left\|v_{0}\right\|}
\end{flalign}
where $V_0$  indicates the current state of UAV, which contains the current position $p_0$ and yaw angle $\xi_0$.  $v_0$ is the current speed of UAV. $N_{cls}$ represents the number of frontiers. $t_{lb} (V_0, V_k)$ and $c_c (V_k)$  represents flight-level factors such as distance, yaw change, and speed change. The calculation method of the rest of $\mathrm{M_{tsp}}$ is consistent with FUEL:
\begin{flalign}
	\begin{aligned}
		&\mathbf{M}_{\mathrm{tsp}}\left(k_{1}, k_{2}\right)=\mathbf{M}_{\mathrm{tsp}}\left(k_{2}, 
		k_{1}\right) \\
		&=t_{\mathrm{lb}}\left({V}_{k_{1}}, {V}_{k_{2}}\right), k_{1}, k_{2} 
		\in\left\{1,2, \cdots, N_{\mathrm{cls}}\right\}
	\end{aligned}
\end{flalign}
\begin{flalign}
	\mathbf{M}_{\mathrm{tsp}}(k, 0)=0, k \in\left\{0,1,2, \cdots, N_{\mathrm{cls}}\right\}
\end{flalign}
\begin{figure}[t]
	\vspace{2mm}
	\centering
	\includegraphics[width=\linewidth, height=5.2cm]{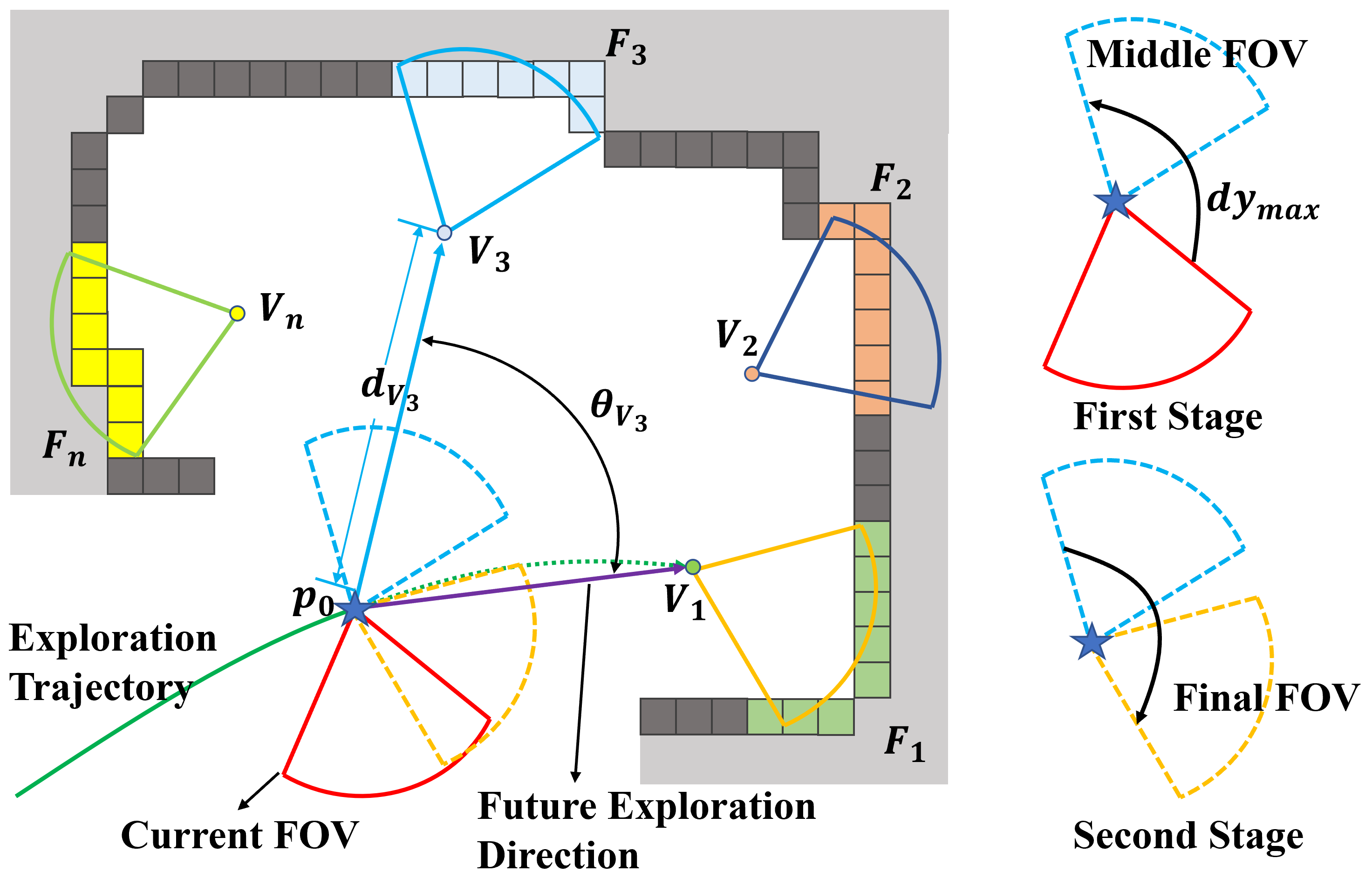}
	\caption{ The proposed Two-Stage heading planning method for the case of multiple viewpoints in a small range: (1) the middle yaw is selected and the corresponding heading planning is conducted in the first stage. (2) the heading planning from the middle yaw to the final yaw is conducted in the second stage.}
	\label{fig:3}
	\vspace{-3mm}
\end{figure}\textbf{}
\subsection{Two-Stage Heading Planning Method}
When the UAV is equipped with limited FOV sensors, heading planning becomes extremely important. An excellent heading planning result can enable the UAV to explore more areas at the same time. Through a large number of experiments, we observe that there are often multiple viewpoints in a small range. If we can make one planning task that can explore multiple frontiers by the heading level planning in the process of flying to a viewpoint of FIS, it will improve the efficiency of exploration. Based on this, this paper designs a two-stage heading planning method to cover more frontiers in an exploration journey, as shown in Fig.\ref{fig:3}, and its main process is described in Algorithm 1, where $V_{n}$ and $X_{0}$ are next target viewpoint and current motion state respectively.

\floatname{algorithm}{Algorithm}  
\renewcommand{\algorithmicrequire}{\textbf{Input:}}  
\renewcommand{\algorithmicensure}{\textbf{Output:}}  

\begin{algorithm}  
    \caption{Two-Stage Heading Planning Method}  
    \begin{algorithmic}[1]
        \Require VPs$(V_1,V_2,…,V_k), V_n (p_n,\xi_n), X_0 (V_0,v_0,a_0)$ 
        \Ensure Heading Trajectory $Y$ 
        \State $N_{v} \gets \textbf{ViewpointsInLocal}$(VPs)
        \State \textbf{if} $N_{v} > 1$ \textbf{then}
	    \State $\quad \xi_{m} \gets \textbf{FindMiddleYaw}$(VPs)  
	    \State $\quad T_{1},T_{2} \gets \textbf{CalculateTwoMinTime}(\xi_{0},\xi_{m},\xi_{n})$   
	    \State $\quad T_{min} \gets \tau \cdot(T_{1}+T_{2}), R \gets T_{1}/T_{min}$   
	    \State $\quad T_{real} \gets \textbf{TrajectoryPlanning}(X_0,p_n,T_{min})$
        \State \quad \textbf{if} $T_{real} >= T_{min}$ \textbf{then}
	    \State $\quad \quad Y_1 \gets \textbf{HeadingPlanning}(\xi_{0},\xi_{m},T_{real}*R)$  
	    \State $\quad \quad Y_2 \gets \textbf{HeadingPlanning}(\xi_{m},\xi_{n},T_{real}*(1-R))$  
	    \State \quad \quad \Return{$Y(Y_{1}, Y_{2})$} 
	    \State \textbf{else}
        \State $\quad T_{min} \gets \textbf{CalculateMinTime}(\xi_{0},\xi_{n})$  
        \State $\quad T_{real} \gets \textbf{TrajectoryPlanning}(X_{0},p_{n},T_{min})$  
        \State $\quad Y \gets \textbf{HeadingPlanning}(\xi_{0},\xi_{n},T_{real})$  
        \State \quad \Return{$Y$}  
    \end{algorithmic}  
\end{algorithm}  

At first, we use function \textbf{ViewpointsInLocal()} to calculate the number of viewpoints $V_{k}$ that are less than $d_{thr}$ and intervisible from the current position $p_{0}$ and the angle ${\theta}_{V_{k}}$ between $\overrightarrow{p_0V_{k}}$ and $\overrightarrow{p_0V_{n}}$ is less than 90 degrees (Line 1). And then, if the number of viewpoints is more than 1, we adopt the multiple viewpoints mode (Line 2-10). Otherwise, the normal heading planning method is used (Line 12-15). Next, \textbf{FindMiddleYaw()} calculates the change between the yaw of each viewpoint and the current yaw, and find the yaw angle with the largest change $\xi_m$ (Line 3). Later, according to the geometric relationship between $\xi_m$, the current yaw $\xi_0$ and the yaw $\xi_n$ of the next target viewpoint, the minimum time $T_{min}$ required for the two heading changes is preliminarily calculated by \textbf{CalculateTwoMinTime()} as follow: 
\begin{flalign}
	T_{1}=\frac{\min \left(\left|\xi_{m}-\xi_{0}\right|, 2 
		\pi-\left|\xi_{m}-\xi_{0}\right|\right)}{\dot{\xi}_{\max }}
\end{flalign}
\begin{flalign}
	T_{2}=\frac{\min \left(\left|\xi_{n}-\xi_{m}\right|, 2 
		\pi-\left|\xi_{n}-\xi_{m}\right|\right)}{\dot{\xi}_{\max }}
\end{flalign}
\begin{flalign}
	T_{\min }=\tau \cdot\left(T_{1}+T_{2}\right)
\end{flalign}

And we provide the $T_{min}$ that is regarded as the minimum flight time constraint, current motion state $X_0$ and the position $p_n$ of the next target viewpoint for \textbf{TrajectoryPlanning()} to generate a flight path (Line 4-6). Finally, if the actual flight time $T_{real}$ is more than $T_{min}$, we conduct two heading planning by \textbf{HeadingPlanning()} (Line7-10, 12-15). In this function, we use a uniform B-spline to represent the trajectory of yaw angle $\phi(t)$, which is parameterized by the  N+1 control points $\Phi:=\{\phi_0, ... \phi_n\}$ and knot span $\delta t_{\phi}$. $T$ is the total time of the trajectory. Due to the convex hull property of B-spline, we can optimize the smoothness and dynamic feasibility of the trajectory by solving the problem:
\begin{flalign}
	\begin{aligned}
		\underset{\xi_{c p}}{\arg \min } \gamma_{1} f_{s}+\gamma_{2} 
		&\left(\phi\left(t_{0}\right)-\xi_{0}\right)+\gamma_{3}\left(\phi(T)-\xi_{n}\right) \\
		&+\gamma_{4}\left(f_{\dot{\xi}}+f_{\ddot{\xi}}\right)
	\end{aligned}
\end{flalign}
where $f_s$ represents smoothness. The second and third terms are soft waypoint constraint enforcing $\phi(t)$ to pass through current yaw $\xi_0$ and target yaw $\xi_n$. The last two terms are the soft constraints for the dynamic feasibility of angular velocity and acceleration. $f_s$, $f_{\dot \xi}$ and $f_{\ddot \xi}$ are similar to  \cite{zhou2021fuel}, \cite{zhou2019robust}.

\begin{figure}[t]
	\vspace{2mm}
	\centering
	\includegraphics[width=1.0\linewidth]{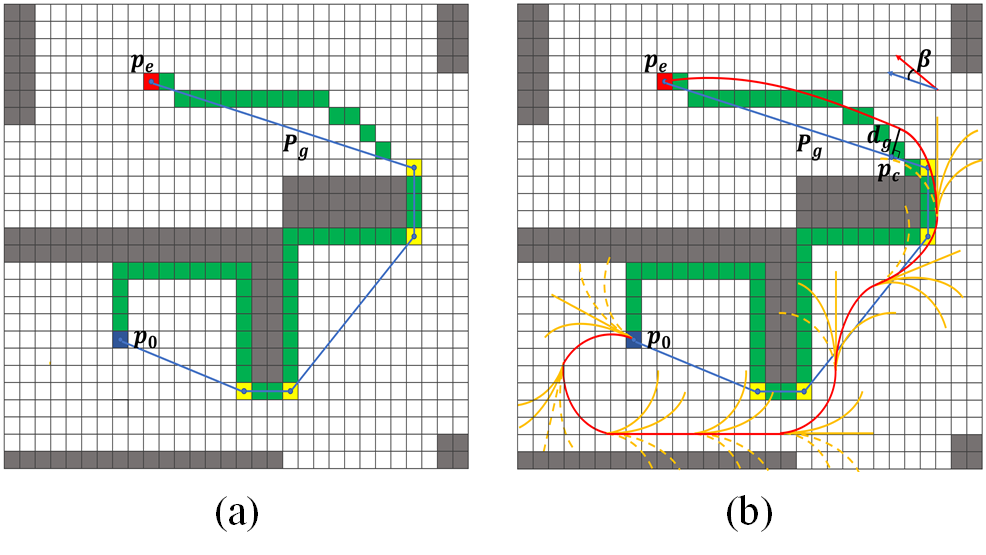}

	\caption{ An illustration of the kinodynamic path Searching method based on the guiding path. Yellow curves indicate the motion primitives. The green grid path is the result of A*. The blue path is the pruning path of the green path. The red curve is the result of the search.}
	\label{fig:4}
	\vspace{-6mm}
\end{figure}\textbf{}
\begin{table*}[!hb]
	\begin{center}
		\caption{EXPLORATION STATISTIC IN THE TWO SCENARIOS}
		\label{tab:1}
		\centering
		\setlength{\tabcolsep}{2.7mm}{
			\renewcommand{\arraystretch}{1.1} {
				\begin{tabular}{cccccccccccccc}
					\toprule
					\multirow{2}{*}{\textbf{Scene}} & \multirow{2}{*}{\textbf{Method}} & \multicolumn{4}{c}{\textbf{Exploration time (s)}}               & \multicolumn{4}{c}{\textbf{Flight distance (m)}}                & \multicolumn{4}{c}{\textbf{Coverage (m3)}}                      \\ \cline{3-14} 
					&                                  & \textbf{Avg}   & \textbf{Std} & \textbf{Max}   & \textbf{Min}   & \textbf{Avg}   & \textbf{Std} & \textbf{Max}   & \textbf{Min}   & \textbf{Avg}   & \textbf{Std} & \textbf{Max}   & \textbf{Min}   \\
					\multirow{4}{*}{Office}         & Proposed                         & \textbf{117.5} & \textbf{5.0} & \textbf{121.8} & \textbf{110.5} & \textbf{163.9} & 12.1         & \textbf{174.4} & \textbf{147.0} & 903.4          & 2.8          & 907.0          & 900.2          \\
					& FUEL                             & 164.8          & 7.7          & 175.6          & 158.3          & 222.4          & 9.1          & 235.3          & 215.0          & \textbf{908.7} & \textbf{0.3} & \textbf{909.0} & \textbf{908.2} \\
					& Aeplanner                        & 338.5          & 11.8         & 353.2          & 324.4          & 200.5          & \textbf{2.3} & 203.2          & 197.5          & 887.8          & 8.1          & 899.0          & 880.5          \\
					& NBVP                             & 637.4          & 162.9        & 839.6          & 440.7          & 305.4          & 73.5         & 390.8          & 211.4          & 860.4          & 79.6         & 950.3          & 756.8          \\ \hline
					\multirow{4}{*}{Outdoor}        & Proposed                         & \textbf{147.3} & \textbf{0.3} & \textbf{147.5} & \textbf{146.9} & \textbf{218.8} & 7.1          & \textbf{226.2} & 209.2          & \textbf{1773}  & \textbf{3.5} & 1777           & \textbf{1769}  \\
					& FUEL                             & 168.9          & 3.7          & 173.4          & 164.3          & 246.3          & \textbf{7.0} & 251.3          & 236.4          & \textbf{1773}  & 4.8          & \textbf{1779}  & 1768           \\
					& Aeplanner                        & 370.4          & 80.5         & 480.8          & 291.4          & 231.7          & 34.7         & 277.1          & \textbf{192.8} & 1705           & 21.7         & 1729           & 1676           \\
					& NBVP                             & 764.0          & 29.0         & 795.1          & 725.3          & 368.7          & 18.1         & 384.2          & 343.4          & 1659           & 103.5        & 1733           & 1513           \\ \bottomrule
				\end{tabular}
		}}
	\end{center}
	\vspace{-3mm}
\end{table*}
\subsection{Guided Kinodynamic Path Searching}
When UAV is in some special scenes, such as searching flight path from inside to outside in a house, if only the conventional kinodynamic path searching is adopted, the search process will take a relatively long time or even failed, which will cause the stop-and-go maneuvers. In order to solve this problem and make the path planning part run more stably and efficiently, this paper adopts the guided kinodynamic path searching as shown in Fig.\ref{fig:4}. A geometric path is generated by A* firstly. And then we prune the path to obtain a guide path $P_g$ that has better guidance. If the distance $d_e$ between $p_c$ and $p_e$ in the guide path is less than 3 or the number of inflection points of the guide path is less than 2, we generate the path by applying Pontryagin’s minimum principle  \cite{mueller2015computationally}:
\begin{flalign}
	p_{\mu}^{*}(t)=\frac{1}{6} \alpha_{\mu} t^{3}+\frac{1}{2} \beta_{\mu} t^{2}+v_{0}+p_{0} \label{equ:11}
\end{flalign}
\begin{flalign}
	\left[\begin{array}{c}
		\alpha_{\mu} \\
		\beta_{\mu}
	\end{array}\right]=\frac{1}{T_{\mu}^{3}}\left[\begin{array}{cc}
		-12 & 6 T_{\mu} \\
		6 T_{\mu} & -2 T_{\mu}^{2}
	\end{array}\right]\left[\begin{array}{c}
		p_{n}-p_{0}-v_{o} T_{\mu} \\
		v_{n}-v_{0}
	\end{array}\right]
\end{flalign}
\begin{flalign}
	\mathcal{J}^{*}\left(T_{\mu}\right)=\sum_{\mu \in\{x, y, z\}}\left(\frac{1}{3} \alpha_{\mu}^{2} 
	T_{\mu}^{3}+\alpha_{\mu} \beta_{\mu} T T_{\mu}^{2}+\beta_{\mu}^{2} T_{\mu}\right)
\end{flalign}
where $v_n$ is the target velocity. The feasible trajectory is 
generated by minimizing the cost $\mathcal{J}^*(T_u)$ of the trajectory. Otherwise, we adopt the guided kinodynamic path searching through a new heuristic function: 
\begin{flalign}
	h_{c}=\lambda_{1} d_{e}+\lambda_{2} d_{g}+\lambda_{3} d_{\theta}, f_{c}=g_{c}+h_{c}
\end{flalign}
where $d_e$ is the distance between $p_c$ and $p_e$ in the guiding path, which is used to improve the efficiency of the search process.  $d_g$ is responsible for constraining the path searching to search in the vicinity of the guiding path. $d_{\theta}$ is used to help the method to find a smoother path. We refer the reader to \cite{zhou2021fuel}, \cite{zhou2019robust}, \cite{zhao2021robust} for more details about kinodynamic path searching and for the path optimization.
\subsection{Adaptive Dynamic Planning}
The speed of target point is usually set zero by default, and the cost time of each replanning is dynamic and unknown. Therefore, if a low-frequency replanning strategy is adopted  and the current position is used as the starting point for planning in the actual flight process, it may cause low speed or stop-and-go maneuvers due to too long time in some planning process, and it may also cause the distance between starting point of the new path and the current position of UAV, which will not maintain a stable and high-speed flight. In order to solve this problem, this paper adopts the strategy of adaptive dynamic starting point for exploration replanning inspired by \cite{tordesillas2019faster}.  In the i-th planning, we do not use the current location as the starting point of the planning, but select the location at the time $t_i$ in the future as the starting point of the current planning, and $t_i$ is not constant, but determined according to the previous planning result:
\begin{flalign}
	t_{i}=\max \left(\rho \cdot t_{i-1}, t_{\min }\right)
	\label{equ:15}
\end{flalign}
where $t_i, t_{i-1}$ represents the cost time of i-th and i-1-th planning respectively.  $t_{min}$  is the minimum time for one planning. If the planning is successful and the actual planning time is less than $t_i$, update the path after time $t_i$ with the planning result. Otherwise, execute replanning. In addition, to maintain the speed and fluency of the flight, we make a replanning when the duration of the remaining flight path is less than 1s.

\begin{figure}[htpb]
	\vspace{2mm}
	\centering
	\includegraphics[width=1.0\linewidth]{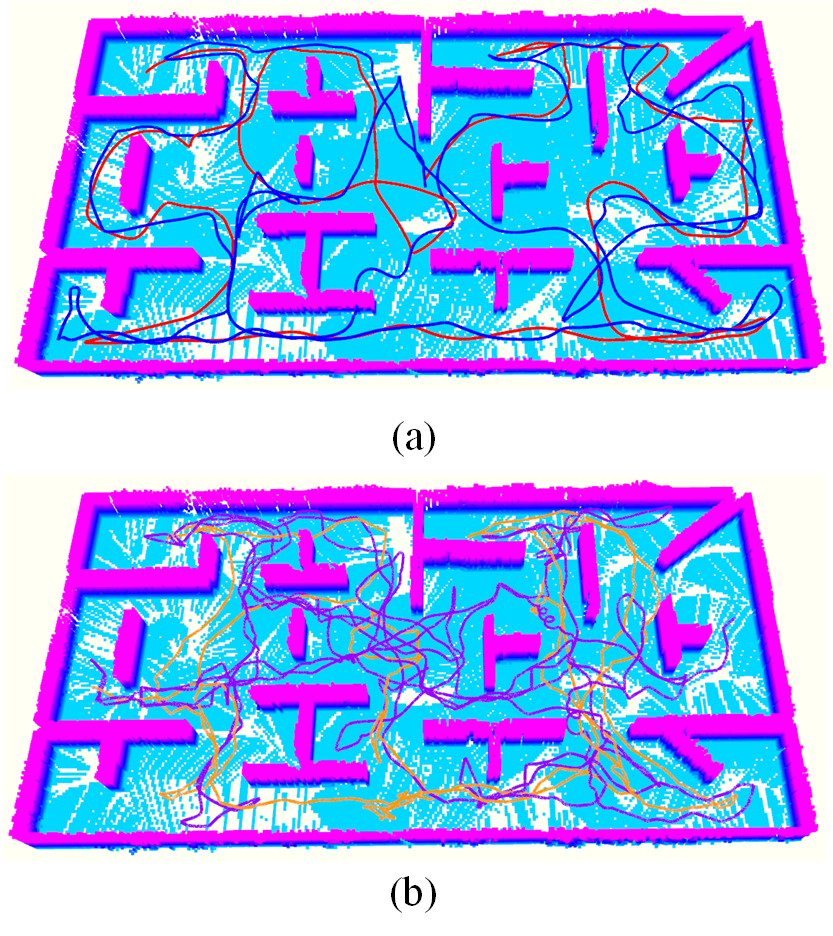}
	\caption{ The flight trajectory comparison of the proposed method (red), FUEL (blue), Aeplanner (orange), and NBVP (purple) in the indoor scene.}
	\label{fig:5}
	\vspace{-3mm}
\end{figure}\textbf{}
\begin{figure}[htpb]
	\centering
	\includegraphics[width=1.0\linewidth]{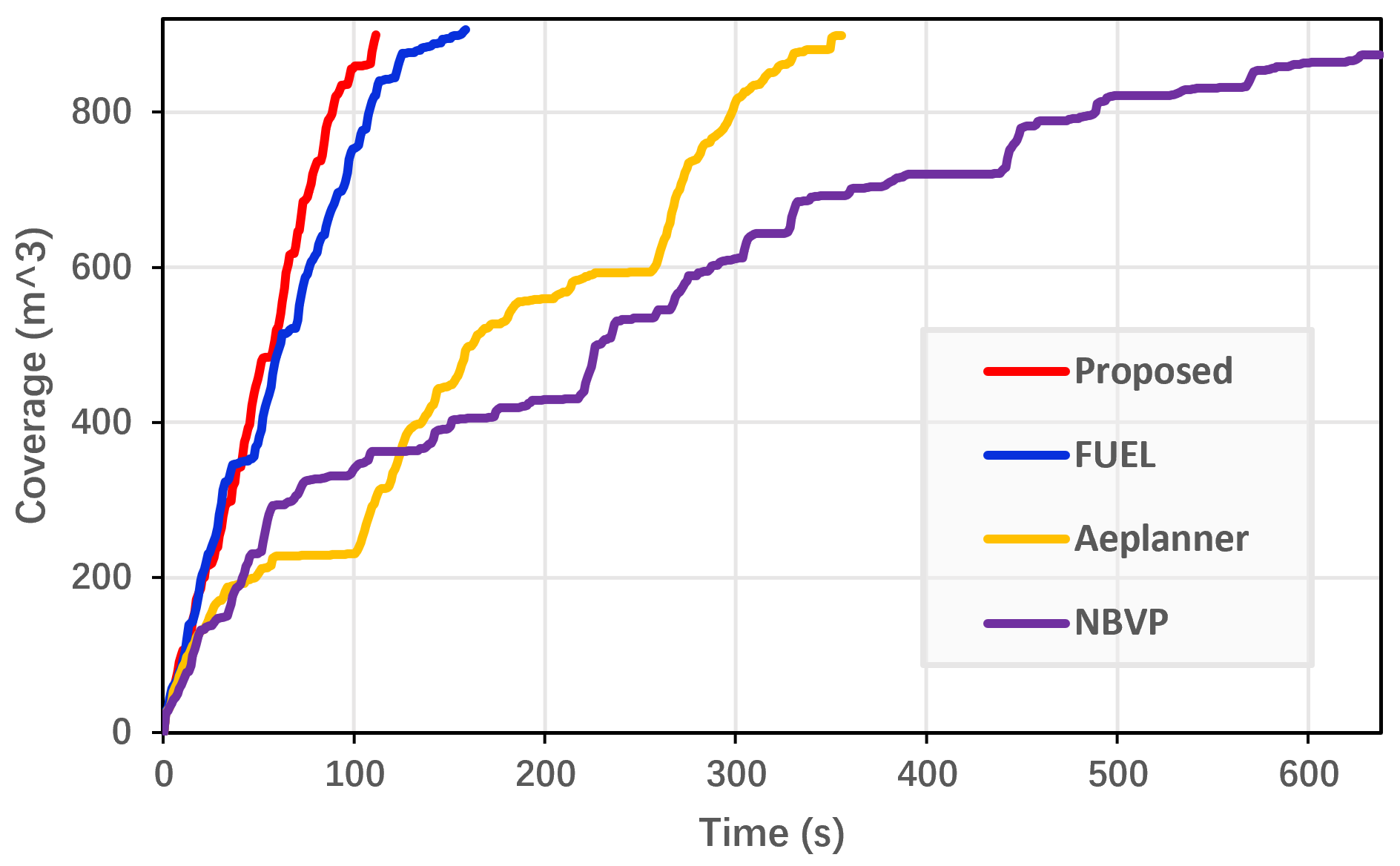}
	\caption{The exploration progress of four methods in the indoor scene.}
	\label{fig:6}
	\vspace{-3mm}
\end{figure}\textbf{}
\section{Experimental Results} 
\label{sec:experiment}
\subsection{Implementation Details}
We set $w_c = 1.5, w_b = 0.3$ and $w_f = 0.3$ in Equ. \ref{equ:2} for for global planning. For motion planning, we set $\tau = 1.3$, $\lambda_1 = 30$, $\lambda_2 = 80$, $\lambda_3 = 80$. In Equ. \ref{equ:15}, we use $t_{min} = 0.1$ and $\rho = 1.3$. The other parameters are consistent with FUEL.
\subsection{Benchmark comparisons}
In the simulation experiment, we compare the proposed method with three state-of-the-art methods in different environments. The three methods are FUEL \cite{zhou2021fuel}, Aeplanner \cite{selin2019efficient}, and NBVP \cite{bircher2016receding}. We all adopt its open-source implementation. And in both scenarios, each method is run 3 times with the same initial configuration. In addition, it should be noted that the dynamic limits we used in the experiment are $v_{max} = 2.0  m/s$ and $\dot \xi_{max} = 1.0 rad/s$ for each method. The FOVs of the sensors are set as [ $80 \times  60$] deg with a maximum range of 4.5 m. And we test these methods on a computer with Inter Core i9-9900K@ 3.6GHz, 64GB memory, and ROS Melodic.

\subsubsection{Office Scenario. }Firstly, we compare the exploration efficiency of the four methods in the office environment, and the scene range is  $30 \times 16 \times 2 m^3$. The experimental results are shown in Fig.\ref{fig:5}, \ref{fig:6}, and Tab.\ref{tab:1}. The experimental results show that NBVP takes the longest time and flight distance, and its exploration efficiency is also unstable. Aeplanner is an improved method of NBVP, its efficiency has been improved compared with the former method because it combines the former with frontier exploration to improve the exploration efficiency. Due to the efficient global coverage path and minimum-time flight path, the proposed method and FUEL have obvious advantages over the above two methods. Not only the actual flight path is smoother, but also the time cost and the length of the flight path are less under the condition of ensuring a high coverage rate. At the same time, the proposed method achieves more efficient exploration efficiency than FUEL due to the fewer back-forth-maneuvers and more stable planning strategy. Compared with FUEL, the average exploration time and flight distance of the proposed are reduced by 28.7\% and 26.3\% respectively, and the exploration ratio tends to be more linear.
\begin{figure}[htpb]
	\vspace{2mm}
	\centering
	\includegraphics[width=1.0\linewidth]{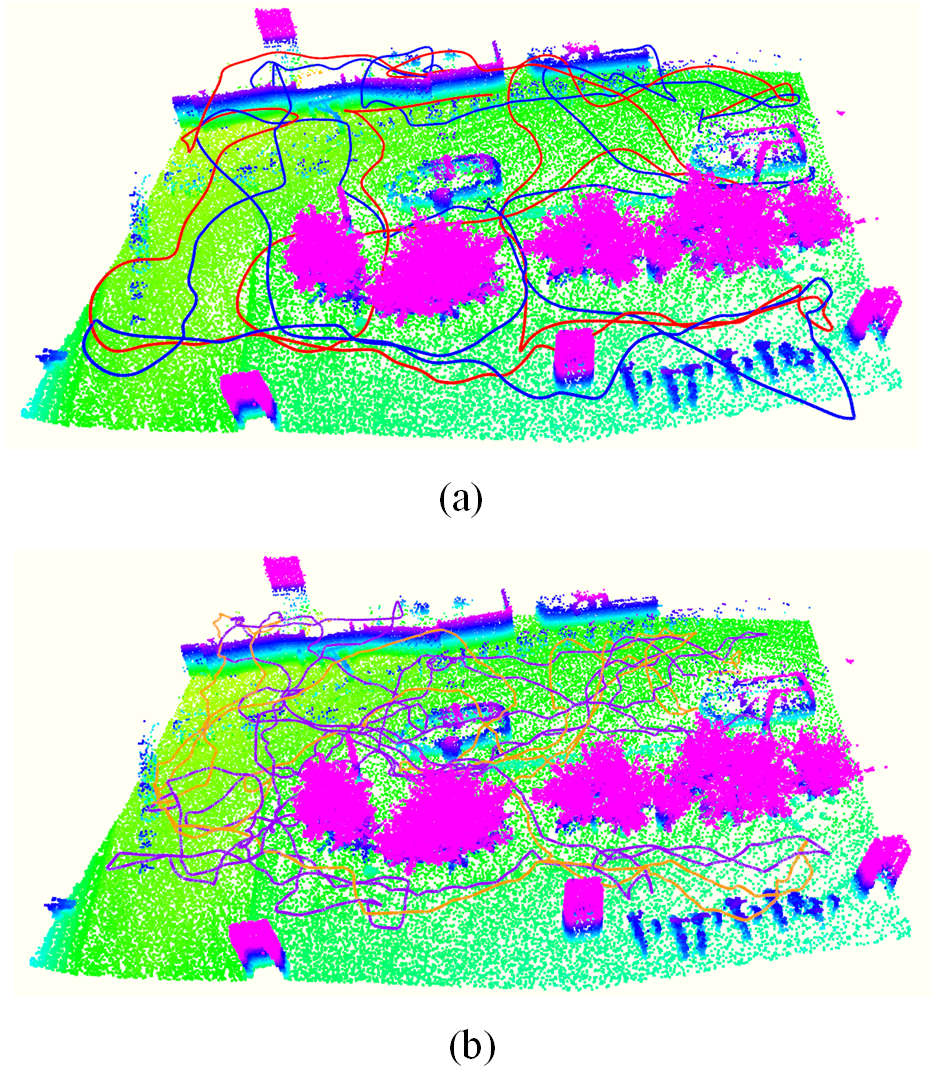}
	\vspace{-6mm}
	\caption{The flight trajectory comparison of the four methods in outdoor scene. The proposed method (red) and FUEL (blue) are in (a). Aeplanner (orange) and NBVP (purple) are in (b).}
	\label{fig:7}
	\vspace{-3mm}
\end{figure}\textbf{}
\begin{figure}[htpb]
	\centering
	\includegraphics[width=1.0\linewidth]{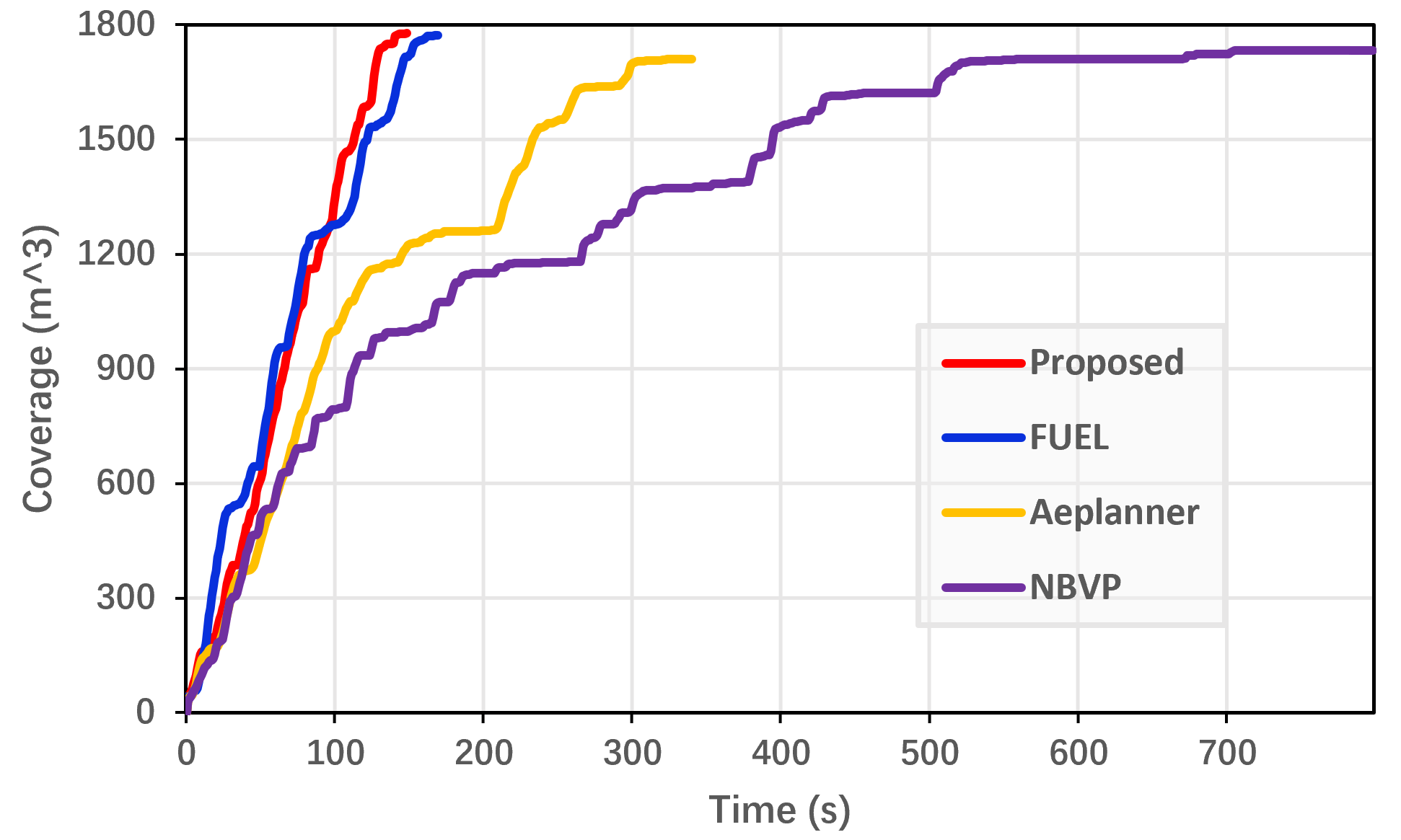}
	\caption{The exploration progress of four methods in the outdoor scenario.}
	\label{fig:8}
	\vspace{-3mm}
\end{figure}\textbf{}
\subsubsection{Outdoor Scenario.} In addition, we also compare the four methods in the outdoor scenario. The scenario contains trees, cars, corridor columns, fences, and other objects, with a range of $20 \times30 \times3 m^3$ . The experimental results are shown in Fig.\ref{fig:7}, \ref{fig:8}, and Tab.\ref{tab:1}. The results show that the exploration time and distance of the four methods are improved compared with the previous scene due to the increase of scene complexity, but the proposed method still maintains obvious advantages in exploration time and distance compared with other methods. Compared with NBVP and Aeplanner, our method achieves the exploration 3-6 times faster on average. And compared with 
FUEL, our method still maintains the advantages of 12.8\% and 11.2\% in exploration time and flight distance respectively.
\begin{figure*}[htpb]
	\vspace{2mm}
	\centering
	\includegraphics[width=.9\linewidth]{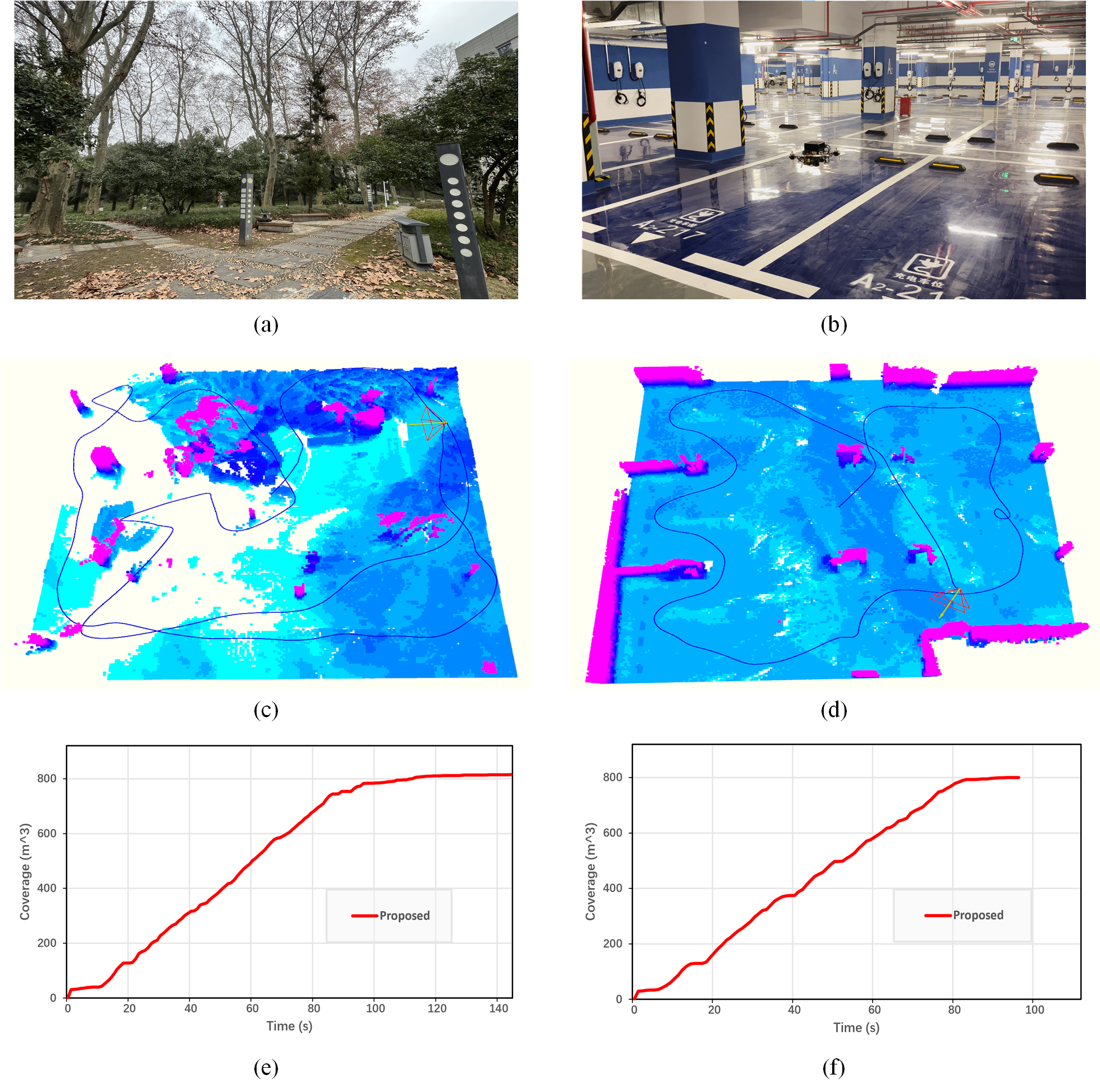}
	\vspace{-6mm}
	\caption{The results of real-world experiments. (a), (c) and (e) are the experiment results in wood. (b), (d) and (f) are the results in an underground park. Videos of the experiments can be found at https://www.youtube.com/watch?v=Rypq6-lIW0U.}
	\label{fig:9}
	\vspace{-3mm}
\end{figure*}\textbf{}
\subsection{Real-world Experiments}
In order to verify the effectiveness of the proposed method, we also conduct two real-world experiments in wood and underground park. In the experiments, we set dynamic limits as $v_{max} = 1.0 m/s$, $a_{max} = 1.0 m/s^2$ and $\dot \xi_{max} = 1.0 rad/s$. We equipped our UAV with a limited FOV sensor (Intel RealSense Depth Camera D435), and use \cite{VINS} to provide the quadrotor state. All the modules run on an Intel Core i5-1135G7@ 2.40GHz, 16GB memory and ROS Melodic.

At first, to validate our method in a natural scenario, we carry out exploration experiments in a wood. The scenario contains trees, bushes, stone stools, and other objects. We bound the range of the scenario for exploration by a $20 \times20 \times2.1 m^3$  box. The exploration results are shown in subgraph (a), (c) and (e) of Fig. \ref{fig:9}. And the exploration time of the whole process is 144.5 s, and the flight distance is 131.0 m. And it should be noted that we only build map for areas where the position is greater than -0.1 m in the z-axis, but the wood contains a depression area, which causes the blank area in subgraph (c) of Fig.\ref{fig:9}. In addition, to verify our method in the underground scenario, we also conduct exploration experiments in an underground park, which mainly contains walls and pillars. We also bound the exploration space by a $20 \times 20 \times2.1 m^3$ box. The experiment results are shown in subgraph (b), (d) and (f) of Fig. \ref{fig:9}. The exploration time and flight distance of the whole exploration process are 94.3 s and 90.2 m respectively. The above two experiments prove that our method can achieve the exploration task of the target area effectively and safely by using the limited FOV sensor in outdoor natural experiments and indoor environments. We also provide a video demonstration in Fig.\ref{fig:9} for readers to get more details.

\section{Conclusion And Future Work} 
\label{sec:conclusion}
Based on the framework of FUEL, this paper proposes a fast and autonomous exploration method (FAEP) for UAVs equipped with limited FOV sensors. Firstly, this paper designs 
a better frontiers exploration sequence generation method, which not only considers the cost of flight-level (distance, yaw change, and velocity direction change) but also considers the influence of the frontier on global exploration. Secondly, according to the flight state of UAV and FISs, a two-stage heading planning strategy is proposed to cover more frontiers in one flight task. Thirdly, a guided kinodynamic path searching method is designed to achieve efficient and stable operation of the planning part. Finally, adaptive dynamic planning is adopted to increase the stability and fluency of the flight process by selecting the dynamic start point and corresponding replanning strategy. Both Simulation and real-world experiments verify the efficiency of our method.

We also look forward to the next work. Although we have designed a method to quantify the influence of frontiers on global exploration, the method is relatively incomplete due to the use of one ray, which can not obtain accurate results and cause low-speed flight in some special environments. In the future, we will study and design a more efficient exploration value evaluation method.

\section*{ACKNOWLEDGMENT}
\label{sec:acknowledgment}
This work was supported by National Key Research and Development Project of China (Grant No. 2020YFD1100200), the Science and Technology Major Project of Hubei Province under Grant (Grant No. 2021AAA010).

\balance
	
\bibliographystyle{IEEEtran}
\bibliography{rootrefer}
\end{document}